\documentclass[10pt,twocolumn,letterpaper]{article}

\usepackage{cvpr}
\usepackage{times}
\usepackage{graphicx}
\usepackage{amsmath}
\usepackage{amssymb}
\usepackage{svg}
\usepackage{subfig}
\usepackage{algorithm, algpseudocode}
\usepackage{multirow}
\usepackage{enumitem}
\usepackage{ctable}
\usepackage{authblk}
\setitemize{noitemsep,topsep=3pt,parsep=3pt,partopsep=0pt}
\setenumerate{noitemsep,topsep=3pt,parsep=3pt,partopsep=0pt}

\usepackage[pagebackref=true,breaklinks=true,letterpaper=true,colorlinks,bookmarks=false]{hyperref}

 \cvprfinalcopy 

\def\cvprPaperID{1246} 

\ifcvprfinal\fi
\begin{document}

\title{Bag of Tricks for Image Classification with Convolutional Neural Networks}

\author[ ]{
Tong He
}
\author[ ]{
\; Zhi Zhang
}
\author[ ]{
\; Hang Zhang
}
\author[ ]{
\; Zhongyue Zhang
}
\author[ ]{
\; Junyuan Xie
}
\author[ ]{
\; Mu Li
}

\affil[ ]{Amazon Web Services}
\affil[ ]{ 
\tt\small \{htong,zhiz,hzaws,zhongyue,junyuanx,mli\}@amazon.com
}
\renewcommand\Authsep{  } 
\renewcommand\Authands{  }

\maketitle

\begin{abstract}

Much of the recent progress made in image classification research can be credited to training procedure refinements, 
such as changes in data augmentations and optimization methods.
In the literature, however, most refinements are either briefly mentioned as implementation details or only visible in source code.
In this paper, we will examine a collection of such refinements and empirically
evaluate their impact on the final model accuracy through ablation study. We
will show that, by combining these refinements together, we are able to improve various CNN models significantly.
For example, we raise ResNet-50's top-1 validation accuracy from
75.3\% to 79.29\% on ImageNet.
We will also demonstrate that improvement on image classification accuracy leads to better transfer learning performance in
other application domains such as object detection and semantic segmentation.
\end{abstract}

\section{Introduction}

Since the introduction of AlexNet~\cite{krizhevsky2012imagenet} in 2012, deep
convolutional neural networks have become the dominating approach for image
classification. Various new architectures have been proposed since then,
including VGG~\cite{DBLP:journals/corr/SimonyanZ14a}, NiN~\cite{lin2013network},
Inception~\cite{DBLP:journals/corr/ChenPSA17}, ResNet~\cite{he2016deep}, DenseNet~\cite{huang2017densely},
and NASNet~\cite{DBLP:journals/corr/ZophVSL17}. At the same time, we have seen a steady trend of
model accuracy improvement. For example, the top-1 validation accuracy on
ImageNet~\cite{russakovsky2015imagenet} has been raised from 62.5\% (AlexNet) to 82.7\% (NASNet-A).

However, these advancements did not solely come from improved model
architecture. Training procedure refinements, including changes in loss functions, data
preprocessing, and optimization methods also played a major role.
A large number of such refinements has been proposed in the past years, but
has received relatively less attention.
In the literature, most were only briefly mentioned as implementation details while others can only be found in source code.

In this paper, we will examine a collection of training procedure and model architecture refinements that improve
model accuracy but barely change computational complexity.
Many of them are minor ``tricks'' like modifying the stride size of a particular
convolution layer or adjusting learning rate schedule. 
Collectively, however, they make a big difference.
We will evaluate them on multiple
network architectures and datasets and report their impact to the final model accuracy.

\begin{table}[t!]
  \centering
  \begin{tabular}{l|l|l|l}\hline
    Model & FLOPs & top-1 & top-5 \\ \specialrule{.1em}{.05em}{.05em}
    ResNet-50~\cite{he2016deep} & 3.9 G & 75.3 & 92.2 \\
    ResNeXt-50~\cite{xie2017aggregated} & 4.2 G & 77.8 & - \\
    SE-ResNet-50~\cite{hu2017squeeze} & 3.9 G & 76.71 & 93.38 \\
    SE-ResNeXt-50~\cite{hu2017squeeze} & 4.3 G &  78.90 & 94.51  \\
    DenseNet-201~\cite{huang2017densely} & 4.3 G & 77.42 & 93.66 \\
    ResNet-50 + tricks (ours) & 4.3 G & \textbf{79.29} & \textbf{94.63} \\\hline
  \end{tabular}
  \caption{\textbf{Computational costs and validation accuracy of various models.} 
  ResNet, trained with our ``tricks'', is able to outperform newer and improved architectures trained with standard pipeline.}
  \label{tab:resnet-50}
\end{table}

Our empirical evaluation shows that several tricks lead to significant accuracy improvement and combining them together can further boost the model accuracy.
We compare ResNet-50, after applying all tricks, to other related networks in Table~\ref{tab:resnet-50}. 
Note that these tricks raises ResNet-50's top-1 validation accuracy from 75.3\% to 79.29\% on ImageNet.
It also outperforms other newer and improved network architectures, such as SE-ResNeXt-50.
In addition, we show that our approach can generalize to other networks (Inception
V3~\cite{DBLP:journals/corr/ChenPSA17} and MobileNet~\cite{howard2017mobilenets}) and datasets (Place365~\cite{zhou2017places}). We further show that models trained with our tricks bring better transfer learning performance in other application domains such as object detection and semantic segmentation.



\paragraph{Paper Outline.} We first set up a baseline training procedure in
Section~\ref{sec:training-procedure}, and then discuss several tricks that are
useful for efficient training on new hardware in Section~\ref{sec:efficient}. In
Section~\ref{sec:arch} we review three minor model architecture tweaks for ResNet and propose a
new one. Four additional training procedure refinements are then discussed in
Section~\ref{sec:bog}. At last, we study if these more accurate models can help
transfer learning in Section~\ref{sec:transfer-learning}.

Our model implementations and training scripts are publicly available in GluonCV \footnote{\url{https://github.com/dmlc/gluon-cv}}.

\section{Training Procedures}
\label{sec:training-procedure}

The template of training a neural network with mini-batch stochastic gradient
descent is shown in Algorithm~\ref{algo:sgd}. In each iteration, we randomly sample $b$
images to compute the gradients and then update the network parameters. It stops
after $K$ passes through the dataset. 
All functions and hyper-parameters in Algorithm~\ref{algo:sgd} can be implemented in many different ways. 
In this section, we first specify a baseline implementation of
Algorithm~\ref{algo:sgd}.

\begin{algorithm}[t!]
\caption{Train a neural network with mini-batch stochastic gradient descent.}
\label{algo:sgd}
\begin{algorithmic}
\State initialize(net)
\For{$\textrm{epoch}=1, \ldots, K$}
\For{$\textrm{batch}=1, \ldots, \textrm{\#images}/b$}
\State images $\gets$ uniformly random sample $b$ images
\State $X, y\gets$ preprocess(images)
\State $z\gets$ forward(net, $X$)
\State $\ell \gets$ loss($z$, $y$)
\State grad $\gets$ backward($\ell$)
\State update(net, grad)
\EndFor
\EndFor
\end{algorithmic}
\end{algorithm}

\subsection{Baseline Training Procedure}
\label{sec:basel-train-proc}

We follow a widely used implementation~\cite{resnet_torch} of ResNet as our baseline.
The preprocessing pipelines between training and validation are different. During training, we perform the following steps one-by-one:
\begin{enumerate}
\item Randomly sample an image and decode it into 32-bit floating point raw pixel values in $[0, 255]$. 
\item Randomly crop a rectangular region whose aspect ratio is randomly sampled
  in $[3/4, 4/3]$ and area randomly sampled in  $[8\%, 100\%]$, then resize the cropped region into a 224-by-224 square image.
\item Flip horizontally with 0.5 probability.
\item Scale hue, saturation, and brightness with coefficients uniformly drawn from $[0.6, 1.4]$.
\item Add PCA noise with a coefficient sampled from a normal distribution $\mathcal{N}(0, 0.1)$.
\item Normalize RGB channels by subtracting
123.68, 116.779, 103.939 and dividing by 58.393, 57.12, 57.375,
respectively.
\end{enumerate}

During validation, we resize each image's shorter edge to $256$ pixels
while keeping its aspect ratio. Next, we crop out the 224-by-224 region in the
center and normalize RGB channels similar to training.
We do not perform any random augmentations during validation.

The weights of both convolutional and fully-connected layers are initialized with the
Xavier algorithm~\cite{glorot2010understanding}. In particular,
we set the parameter to random values uniformly drawn from $[-a, a]$,
where $a = \sqrt{6/(d_{in} + d_{out})}$. Here $d_{in}$ and $d_{out}$  are the input and output channel sizes,
respectively. All biases are initialized to 0.
For batch normalization layers, $\gamma$ vectors are initialized to 1 and $\beta$ vectors to 0.

Nesterov Accelerated Gradient (NAG) descent~\cite{nesterov1983method} is used for training.
Each model is trained for 120 epochs on 8 Nvidia V100 GPUs with a total batch size of 256.
The learning rate is initialized to $0.1$ and divided by 10 at the
30th, 60th, and 90th epochs.

\subsection{Experiment Results}

We evaluate three CNNs: ResNet-50~\cite{he2016deep}, Inception-V3~\cite{DBLP:journals/corr/ChenPSA17}, and
MobileNet~\cite{howard2017mobilenets}. For Inception-V3 we resize the input images into 299x299.
We use the ISLVRC2012~\cite{russakovsky2015imagenet} dataset, which has 1.3
million images for training and 1000 classes. The validation accuracies are
shown in Table~\ref{tab:baseline}. As can be seen, our ResNet-50 results are slightly better than
the reference results, while our baseline Inception-V3 and MobileNet are slightly
lower in accuracy due to different training procedure.

\begin{table}[t!]
\begin{center}
  \begin{tabular}{l|l|l|l|l}
    \hline
    \multirow{2}{*}{Model}& \multicolumn{2}{c|}{Baseline} & \multicolumn{2}{c}{Reference}\\\cline{2-5}
    & Top-1 & Top-5 & Top-1 & Top-5 \\\specialrule{.1em}{.05em}{.05em}
    ResNet-50~\cite{he2016deep} & 75.87 & 92.70 & 75.3 & 92.2 \\\hline    
    Inception-V3~\cite{DBLP:journals/corr/SzegedyVISW15} & 77.32 & 93.43 & 78.8 & 94.4 \\\hline
    MobileNet~\cite{howard2017mobilenets} & 69.03 & 88.71 & 70.6 & - \\\hline
\end{tabular}   
\end{center}
\caption{\textbf{Validation accuracy of reference implementations and our baseline.} Note that the numbers for Inception V3 are obtained with 299-by-299 input images.}
\label{tab:baseline}
\end{table}

\section{Efficient Training}
\label{sec:efficient}

Hardware, especially GPUs, has been rapidly evolving in recent years.
As a result, the optimal choices for many performance related trade-offs have changed.
For example, it is now more efficient to use lower numerical precision and larger batch sizes during training.
In this section, we review various techniques that enable low precision and large batch training without sacrificing model accuracy.
Some techniques can even improve both accuracy and training speed.


\subsection{Large-batch training}
\label{sec:large-batch-training}


Mini-batch SGD groups multiple samples to a mini-batch to increase parallelism and decrease communication costs.
Using large batch size, however, may slow down the training progress.
For convex problems, convergence rate decreases as batch size increases. Similar empirical results have been reported for neural networks ~\cite{smith2017don}. In other words, for the same number of epochs, training with a large batch size results in a model with degraded validation accuracy compared to the ones trained with smaller batch sizes.

Multiple works~\cite{DBLP:journals/corr/GoyalDGNWKTJH17, jia2018highly} have proposed heuristics to solve this issue. In the following paragraphs, we will examine four heuristics that help scale the batch size up for single machine training.

\paragraph{Linear scaling learning rate.} In mini-batch SGD,  gradient descending is a random process because the examples are randomly selected in each batch. Increasing the batch size does not change the expectation of the stochastic gradient but reduces its variance.  In other words, a large batch size reduces the noise in the gradient, so we may increase the learning rate to make a larger progress along the opposite of the gradient direction. Goyal \etal~\cite{DBLP:journals/corr/GoyalDGNWKTJH17} reports that linearly increasing the learning rate with the batch size works empirically for ResNet-50 training. In particular, if we follow He~\etal~\cite{he2016deep} to choose 0.1 as the initial learning rate for batch size 256, then when changing to a larger batch size $b$, we will increase the initial learning rate to $0.1\times b/256$.

\paragraph{Learning rate warmup.} At the beginning of the training, all parameters are typically random values and therefore far away from the final solution. Using a too large learning rate may result in numerical instability. In the warmup heuristic, we use a small learning rate at the beginning and then switch back to the initial learning rate when the training process is stable~\cite{he2016deep}. Goyal \etal~\cite{DBLP:journals/corr/GoyalDGNWKTJH17} proposes a gradual warmup strategy that increases the learning rate from 0 to the initial learning rate linearly. In other words, assume we will use the first $m$ batches (e.g. 5 data epochs) to warm up, and the initial learning rate is $\eta$, then at batch $i$, $1\le i \le m$, we will set the learning rate to be $i\eta/m$.

\paragraph{Zero $\gamma$.} A ResNet network consists of multiple residual blocks, each block consists of several convolutional layers. Given input $x$, assume $\textrm{block}(x)$ is the output for the last layer in the block, this residual block then outputs $x+\textrm{block}(x)$. Note that the last layer of a block could be a batch normalization (BN) layer. The BN layer first standardizes its input, denoted by $\hat x$, and then performs a scale transformation $\gamma \hat x + \beta$. Both $\gamma$ and $\beta$ are learnable parameters whose elements are initialized to 1s and 0s, respectively. In the zero $\gamma$ initialization heuristic, we initialize $\gamma=0$ for all BN layers that sit at the end of a residual block. Therefore, all residual blocks just return their inputs, mimics network that has less number of layers and is easier to train at the initial stage.

\paragraph{No bias decay.} The weight decay is often applied to all learnable parameters including both weights and bias. It's equivalent to applying an L2 regularization to all parameters to drive their values towards 0. As pointed out by Jia \etal~\cite{jia2018highly}, however, it's  recommended to only apply the regularization to weights to avoid overfitting. The no bias decay heuristic follows this recommendation, it only applies the weight decay to the weights in convolution and fully-connected layers. Other parameters, including the biases and $\gamma$ and $\beta$ in BN layers, are left unregularized.

Note that LARS~\cite{ginsburg2018large} offers layer-wise adaptive learning rate and is reported to be effective for extremely large batch sizes (beyond 16K). While in this paper we limit ourselves to methods that are sufficient for single machine training, in which case a batch size no more than 2K often leads to good system efficiency.

\subsection{Low-precision training}

Neural networks are commonly trained with 32-bit floating point (FP32) precision. That is, all numbers are stored in FP32 format and both inputs and outputs of arithmetic operations are FP32 numbers as well. New hardware, however, may have enhanced arithmetic logic unit for lower precision data types. For example, the previously mentioned Nvidia V100 offers 14 TFLOPS in FP32 but over 100 TFLOPS in FP16. As in Table~\ref{tab:efficient}, the overall training speed is accelerated by 2 to 3 times after switching from FP32 to FP16 on V100.

Despite the performance benefit, a reduced precision has a narrower range that makes results more likely to be out-of-range and then disturb the training progress. Micikevicius~\etal~\cite{micikevicius2017mixed} proposes to store all parameters and activations in FP16 and use FP16 to compute gradients. At the same time, all parameters have an copy in FP32 for parameter updating. In addition, multiplying a scalar to the loss to better align the range of the gradient into FP16 is also a practical solution.

\subsection{Experiment Results}

The evaluation results for ResNet-50 are shown in
Table~\ref{tab:efficient}. Compared to the baseline with batch size 256 and
FP32, using a larger 1024 batch size and FP16 reduces the training time for ResNet-50 
from 13.3-min per epoch to 4.4-min per epoch. In addition, by stacking all heuristics
for large-batch training, the model trained with 1024 batch size and FP16 even
slightly increased 0.5\% top-1 accuracy compared to the baseline model.

The ablation study of all heuristics is shown in
Table~\ref{tab:efficient-breakdown}. Increasing batch size from
256 to 1024 by linear scaling learning rate alone leads to a 0.9\% decrease
of the top-1 accuracy while stacking the rest three heuristics bridges the gap.
Switching from FP32 to FP16 at the end of training does not affect the accuracy.

\begin{table*}
\begin{center}
  \begin{tabular}{l|l|l|l|l|l|l}
  \hline
  \multirow{2}{*}{Model}& \multicolumn{3}{c|}{Efficient} & \multicolumn{3}{c}{Baseline} \\\cline{2-7}
  &  Time/epoch & Top-1 & Top-5 &  Time/epoch & Top-1 & Top-5 \\\specialrule{.1em}{.05em}{.05em}
  ResNet-50 & \textbf{4.4 min} & \textbf{76.21} & \textbf{92.97} & 13.3 min & 75.87 & 92.70 \\\hline
  Inception-V3 & \textbf{8 min} & \textbf{77.50} & \textbf{93.60} & 19.8 min & 77.32 & 93.43 \\\hline
  MobileNet & \textbf{3.7 min} & \textbf{71.90} & \textbf{90.47} & 6.2 min & 69.03 & 88.71 \\\hline
  \end{tabular}
\end{center}
\caption{Comparison of the training time and validation accuracy for ResNet-50 between the baseline (BS=256 with FP32) and a more hardware efficient setting (BS=1024 with FP16).}
\label{tab:efficient}
\end{table*}

\begin{table}
\begin{center}
  \begin{tabular}{l|l|l|l|l}
    \hline
    \multirow{2}{*}{Heuristic} & \multicolumn{2}{c|}{BS=256} & \multicolumn{2}{c}{BS=1024}\\\cline{2-5}
     & Top-1 & Top-5 & Top-1 & Top-5 \\\specialrule{.1em}{.05em}{.05em}
    Linear scaling & 75.87 & 92.70 & 75.17 & 92.54 \\
    + LR warmup & 76.03 & 92.81 & 75.93 & 92.84 \\
    + Zero $\gamma$ & 76.19 & 93.03 & 76.37 & 92.96 \\
    + No bias decay & 76.16 & 92.97 & 76.03 & 92.86\\
    + FP16 & 76.15 & 93.09 & 76.21 & 92.97 \\\hline
\end{tabular}
\end{center}
\caption{The breakdown effect for each effective training heuristic on ResNet-50.}
\label{tab:efficient-breakdown}
\end{table}

\begin{figure}[t!]
  \centering
  \includegraphics[scale=.55]{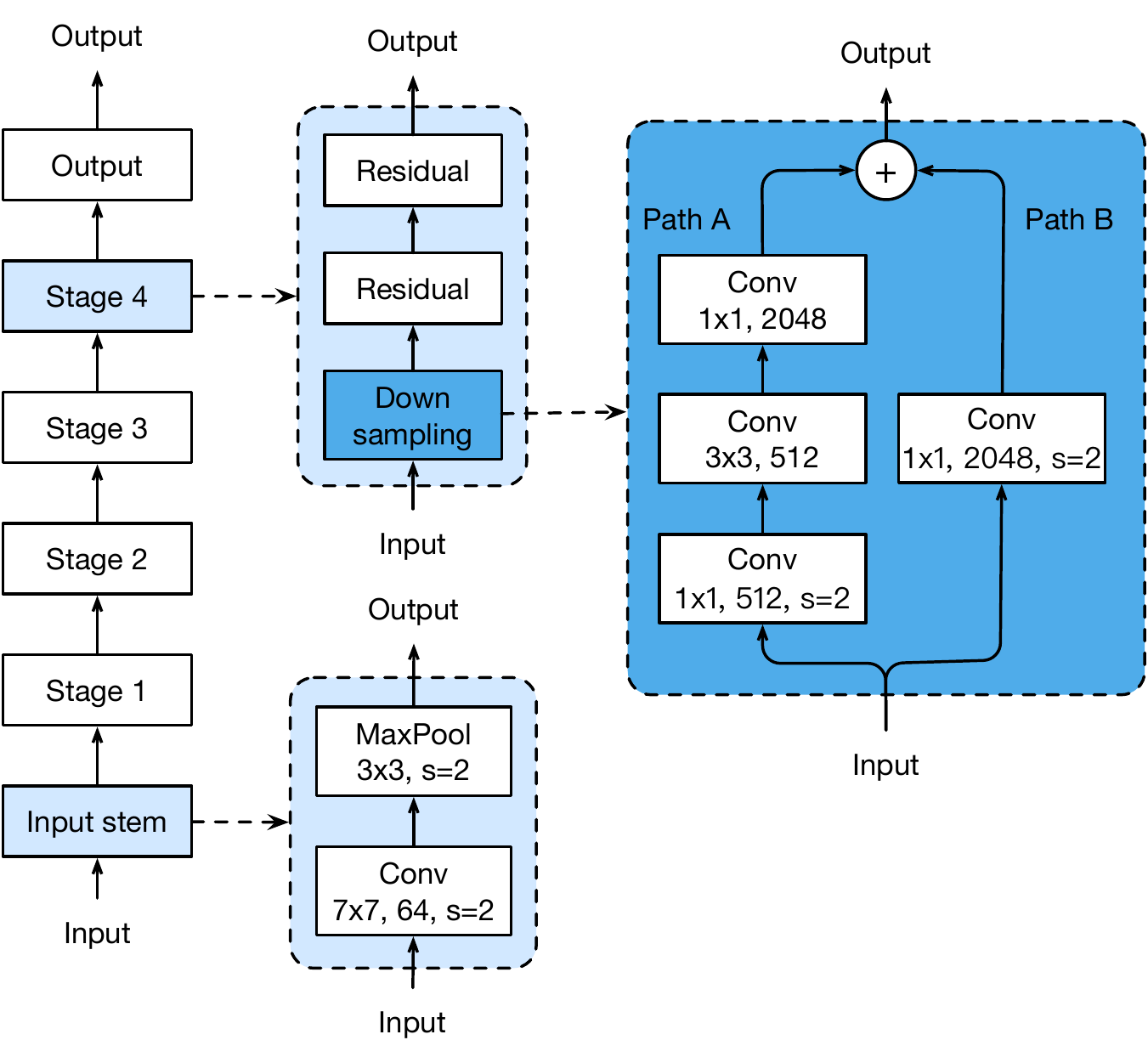}
  \caption{The architecture of ResNet-50. The convolution kernel size, output
    channel size and stride size (default is 1)
    are illustrated, similar for pooling layers.}
  \label{fig:resnet-a}
\end{figure}

\section{Model Tweaks}
\label{sec:arch}

A model tweak is a minor adjustment to the network architecture, such as changing
the stride of a particular convolution layer. Such a tweak often barely changes
the computational complexity but might have a non-negligible effect on
the model accuracy. In this section, we will use ResNet as an example to
investigate the effects of model tweaks.

\subsection{ResNet Architecture}

We will briefly present the ResNet architecture, especially its modules related
to the model tweaks. For detailed information please refer to He~\etal~\cite{he2016deep}.
A ResNet network consists of an input stem, four subsequent stages and a final output layer,
which is illustrated in Figure~\ref{fig:resnet-a}. The input stem has a $7\times 7$
convolution with an output channel of 64 and a stride of 2, followed by a
$3\times 3$ max pooling layer also with a stride of 2. The input stem reduces the
input width and height by 4 times and increases its channel size to 64.

Starting from stage 2, each stage begins with a downsampling block, which is then followed by several
residual blocks. In the downsampling block, there are path A and path B. Path A
has three convolutions, whose kernel sizes are $1\times1$, $3\times 3$ and
$1\times 1$, respectively. The first convolution has a stride of 2 to halve the
input width and height, and the last convolution's output channel is 4 times
larger than the previous two, which is called the bottleneck structure. Path B
uses a $1\times 1$ convolution with a stride of 2 to transform the input shape
to be the output shape of path A, so we can sum outputs of both paths to
obtain the output of the downsampling block. A residual block is similar to a
downsampling block except for only using convolutions with a stride of 1.

One can vary the number of residual blocks in each stage to
obtain different ResNet models, such as ResNet-50 and ResNet-152, where the
number presents the number of convolutional layers in the network.

\begin{figure}[t!]
  \centering
  \subfloat[{\fontsize{7}{0}\selectfont ResNet-B}\label{fig:resnet-b}]{
    \includegraphics[scale=.55]{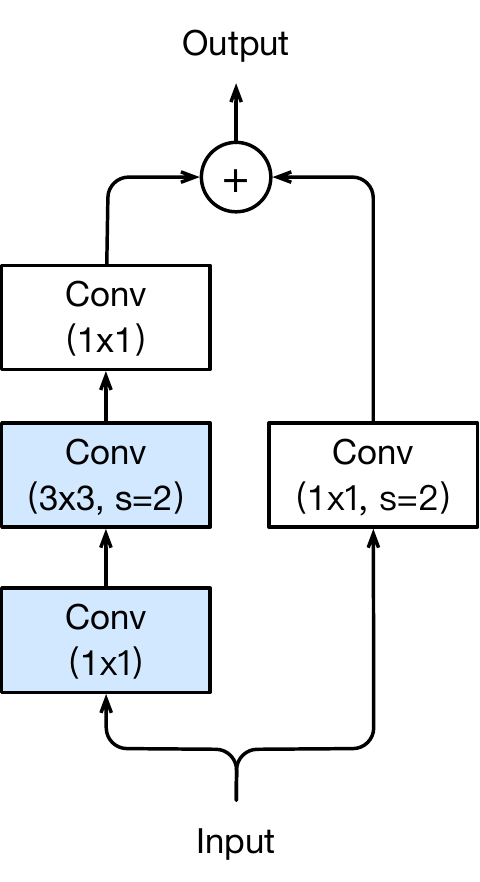}
  }\hfill%
  \subfloat[{\fontsize{7}{0}\selectfont ResNet-C}\label{fig:resnet-c}]{
    \includegraphics[scale=.55]{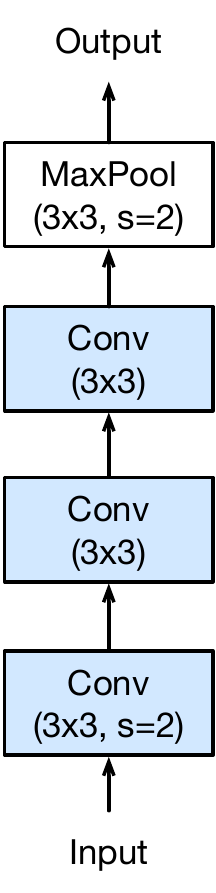}
  }\hfill%
  \subfloat[{\fontsize{7}{0}\selectfont ResNet-D}\label{fig:resnet-d}]{
    \includegraphics[scale=.55]{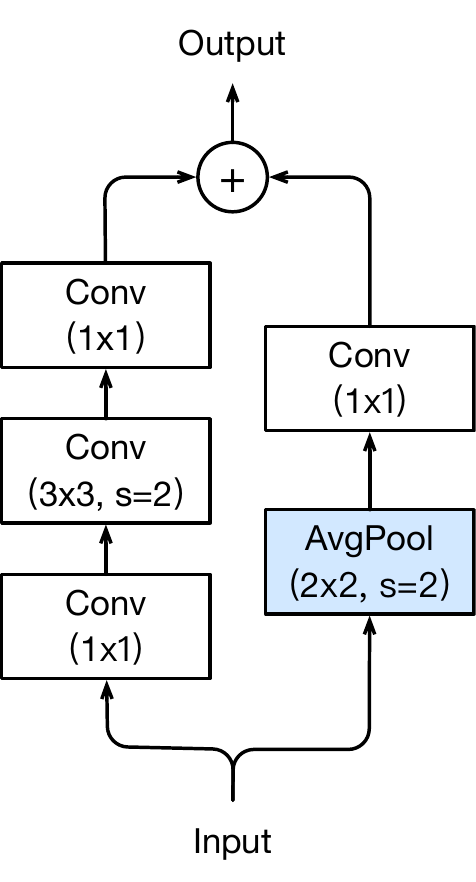}
  }
  \caption{Three ResNet tweaks. ResNet-B modifies the downsampling block of Resnet. ResNet-C further modifies the input stem. On top of that, ResNet-D again modifies the downsampling block.}
  \label{fig:resnet-tweaks}
\end{figure}

\subsection{ResNet Tweaks}

Next, we revisit two popular ResNet tweaks, we call them ResNet-B and
ResNet-C, respectively. 
We propose a new model tweak ResNet-D afterwards. 

\paragraph{ResNet-B.} This tweak first appeared in a Torch implementation of
ResNet~\cite{resnet_torch} and then adopted by multiple works~\cite{DBLP:journals/corr/GoyalDGNWKTJH17, hu2017squeeze,
  xie2017aggregated}. It changes the downsampling block of ResNet. The
observation is that the convolution in path A ignores three-quarters of the
input feature map because it uses a kernel size $1\times 1$ with a stride of
2. ResNet-B switches the strides size of the first two convolutions in
path A, as shown in Figure~\ref{fig:resnet-b}, so no information is ignored.
Because the second convolution has a kernel size $3\times 3$, the output
shape of path A remains unchanged.

\paragraph{ResNet-C.} This tweak was proposed in
Inception-v2~\cite{DBLP:journals/corr/SzegedyVISW15} originally, and it can be
found on the implementations of other models, such as
SENet~\cite{hu2017squeeze}, PSPNet~\cite{zhao2017pyramid}, DeepLabV3~\cite{DBLP:journals/corr/ChenPSA17},
and ShuffleNetV2~\cite{shufflenetv2}. The observation is that the computational
cost of a convolution is quadratic to the kernel width or height. A $7\times 7$
convolution is 5.4 times more expensive than a $3\times 3$ convolution. So this
tweak replacing the $7\times 7$ convolution in the input stem with three
conservative $3\times 3$ convolutions, which is shown in
Figure~\ref{fig:resnet-c}, with the first and second convolutions have their output channel of 32
and a stride of 2, while the last convolution uses a 64 output channel.

\paragraph{ResNet-D.} Inspired by ResNet-B, we note that the $1\times 1$
convolution in the path B of the downsampling block also ignores 3/4 of
input feature maps, we would like to modify it so no information will be
ignored. Empirically, we found adding a $2\times 2$ average pooling layer
with a stride of 2 before the convolution, whose stride is changed to 1, works
well in practice and impacts the computational cost little. This tweak is
illustrated in Figure~\ref{fig:resnet-d}.

\subsection{Experiment Results}

\begin{table}
\begin{center}
\begin{tabular}{l|c|c|c|c}
\hline
Model    &  \#params & FLOPs & Top-1 & Top-5 \\ \specialrule{.1em}{.05em}{.05em}
ResNet-50  & 25 M & \textbf{3.8 G} &  76.21 & 92.97   \\\hline
ResNet-50-B  & 25 M & 4.1 G & 76.66 & 93.28   \\\hline
ResNet-50-C  & 25 M & 4.3 G & 76.87 & 93.48   \\\hline
ResNet-50-D  & 25 M & 4.3 G & \textbf{77.16} & \textbf{93.52} \\ \hline
\end{tabular}
\end{center}
\caption{Compare ResNet-50 with three model tweaks on model size, FLOPs
  and ImageNet validation accuracy.}
\label{tab:resnet-variants}
\end{table}

We evaluate ResNet-50 with the three tweaks and settings
described in Section~\ref{sec:efficient}, namely the batch size is 1024 and
precision is FP16. The results are shown in Table~\ref{tab:resnet-variants}.

Suggested by the results, ResNet-B receives more information in path A of the downsampling
blocks and improves validation accuracy by around $0.5\%$ compared to
ResNet-50. Replacing the $7\times 7$ convolution with three $3\times 3$ ones gives
another $0.2\%$ improvement. Taking more information in path B of the
downsampling blocks improves the validation accuracy by another $0.3\%$. In
total, ResNet-50-D improves ResNet-50 by $1\%$.

On the other hand, these four models have the same model size. ResNet-D has the largest
computational cost, but its difference compared to ResNet-50 is within 15\% in
terms of floating point operations. In practice, we observed ResNet-50-D is only 3\% slower in
training throughput compared to ResNet-50.

\section{Training Refinements}\label{sec:bog}

In this section, we will describe four training refinements that aim to further
improve the model accuracy.

\subsection{Cosine Learning Rate Decay}

Learning rate adjustment is crucial to the training. After the learning
rate warmup described in Section~\ref{sec:large-batch-training}, we typically
steadily decrease the value from the initial learning rate. The widely used
strategy is exponentially decaying the learning rate. He~\etal~\cite{he2016deep} decreases rate at 0.1
for every 30 epochs, we call it ``step decay''.
Szegedy~\etal~\cite{DBLP:journals/corr/SzegedyVISW15} decreases rate at 0.94 for every two epochs.

In contrast to it, Loshchilov~\etal~\cite{DBLP:journals/corr/LoshchilovH16a} propose a cosine
annealing strategy. An simplified version is decreasing the learning rate from
the initial value to 0 by following the cosine function. Assume the total number
of batches is $T$ (the warmup stage is ignored), then at batch $t$, the learning
rate $\eta_t$ is computed as:
\begin{equation}
\eta_t = \frac 12 \left(1 + \textrm{cos}\left(\frac{t\pi}{T}\right)\right)\eta,
\end{equation}
where $\eta$ is the initial learning rate. We call this scheduling as ``cosine''
decay.

\begin{figure}[t!]
  \centering
  \subfloat[Learning Rate Schedule\label{fig:lr-schedule}]{%
    \includegraphics[scale=0.15]{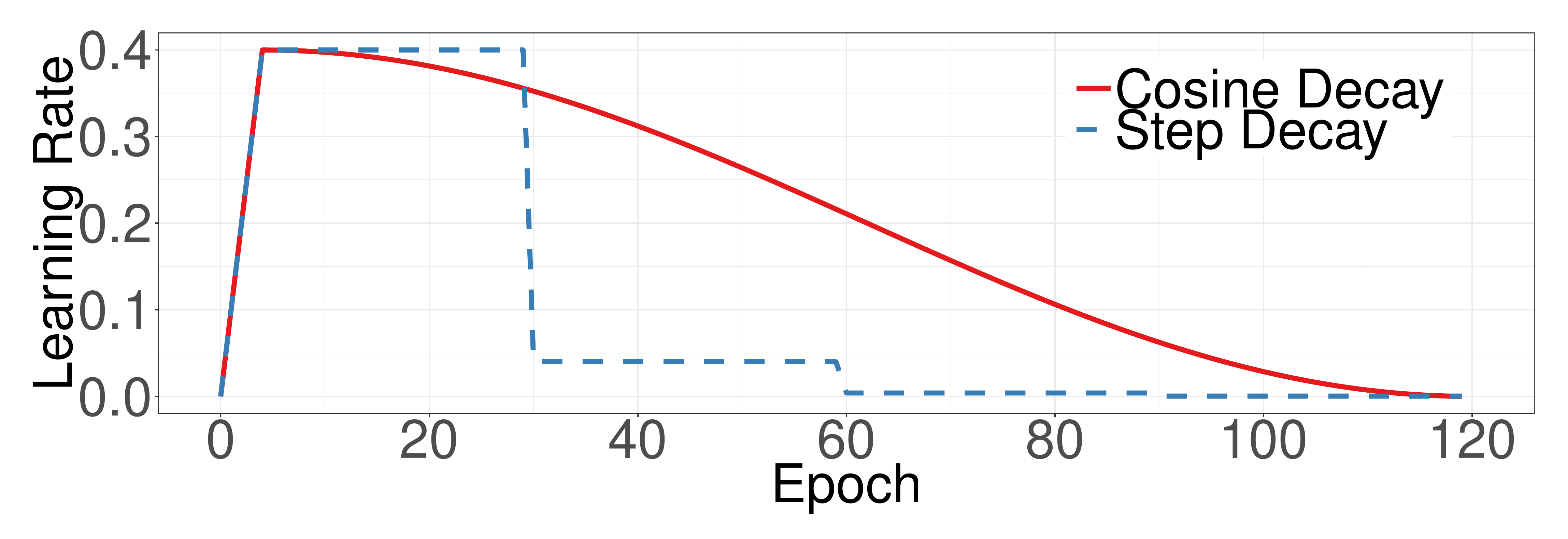}
    }\vfill%
  \subfloat[Validation Accuracy\label{fig:lr-curve}]{%
    \includegraphics[scale=0.15]{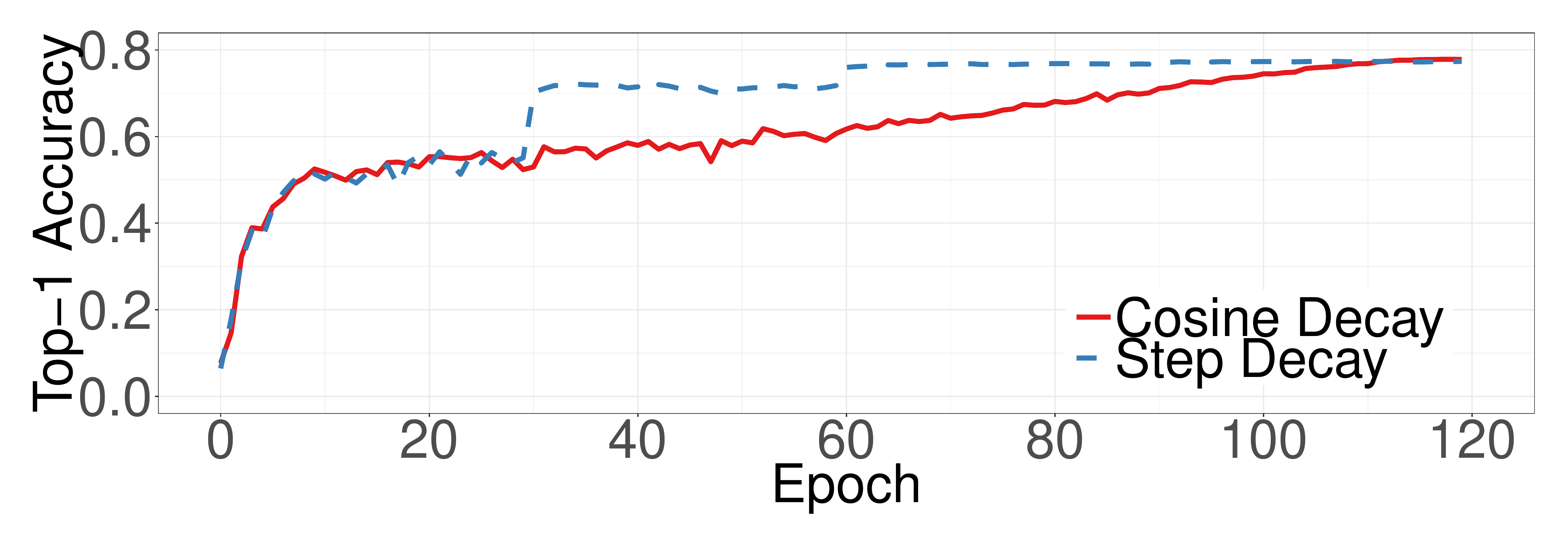}
    }%
  \caption{Visualization of learning rate schedules with warm-up. Top: cosine and step schedules for batch size 1024. Bottom: Top-1 validation accuracy curve with regard to the two schedules.}
  \label{fig:learning-rate-curve}
\end{figure}

The comparison between step decay and cosine decay are illustrated in
Figure~\ref{fig:lr-schedule}. As can be seen, the cosine decay decreases the
learning rate slowly at the beginning, and then becomes almost linear decreasing
in the middle, and slows down again at the end. Compared to the step decay, the
cosine decay starts to decay the learning since the beginning but remains 
large until step decay reduces the learning rate by 10x, which
potentially improves the training progress.

\subsection{Label Smoothing}

The last layer of a image classification network is often a fully-connected
layer with a hidden size being equal to the number of labels, denote by $K$, to
output the predicted confidence scores. Given an image, denote by $z_i$ the
predicted score for class $i$. These scores can be normalized by the softmax
operator to obtain predicted probabilities. Denote by $q$ the output of the
softmax operator $q=\textrm{softmax}(z)$, the probability for
class $i$, $q_i$, can be computed by:
\begin{equation}
q_i = \frac{\exp(z_i)}{\sum_{j=1}^K \exp(z_j)}.
\end{equation}
It's easy to see $q_i > 0$ and $\sum_{i=1}^K q_i = 1$, so $q$ is a valid
probability distribution.

On the other hand, assume the true label of this image is $y$, we can
construct a truth probability distribution to be $p_i = 1$ if $i=y$ and 0 otherwise. During
training, we minimize the negative cross entropy loss

\begin{equation}
\ell(p, q) = - \sum_{i=1}^K q_i \log p_i
\end{equation}
to update model parameters to make these two
probability distributions similar to each other. In particular, by the way how $p$ is
constructed, we know $\ell(p, q) = - \log p_y = - {z_y} + \log\left(\sum_{i=1}^K
    \exp(z_i)\right)$. The optimal solution is $z_y^*=\inf$ while
keeping others small enough. In other words, it encourages the output scores
dramatically distinctive which potentially leads to overfitting.

The idea of label smoothing was first proposed to train
Inception-v2~\cite{DBLP:journals/corr/SzegedyVISW15}. It changes the
construction of the true probability to
\begin{equation}
q_i =
\begin{cases}
  1-\varepsilon & \quad\textrm{if } i = y, \\
   \varepsilon / (K-1) & \quad\textrm{otherwise,}\\
\end{cases}
\end{equation}

where $\varepsilon$ is a small constant. Now the optimal solution becomes

\begin{equation}
z_i^* =
\begin{cases}
  \log( (K-1)(1-\varepsilon)/\varepsilon) + \alpha & \quad\textrm{if } i = y, \\
   \alpha & \quad\textrm{otherwise,}\\
\end{cases}
\end{equation}

where $\alpha$ can be an arbitrary real number. This encourages a finite output from the 
fully-connected layer and can generalize better.

\begin{figure}[t!]
  \centering
  \subfloat[Theoretical gap\label{fig:label_smoothing}]{%
    \includegraphics[scale=0.15]{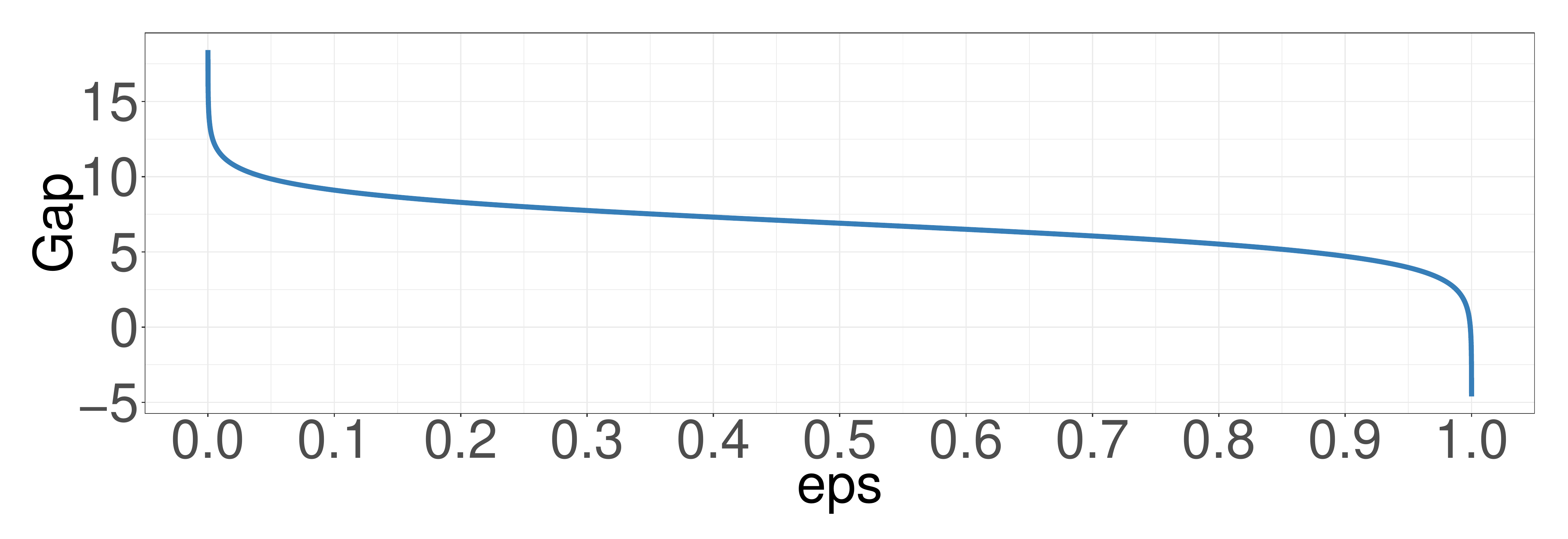}
    }\vfill%
  \subfloat[Empirical gap from ImageNet validation set\label{fig:label_smoothing_density}]{%
    \includegraphics[scale=0.15]{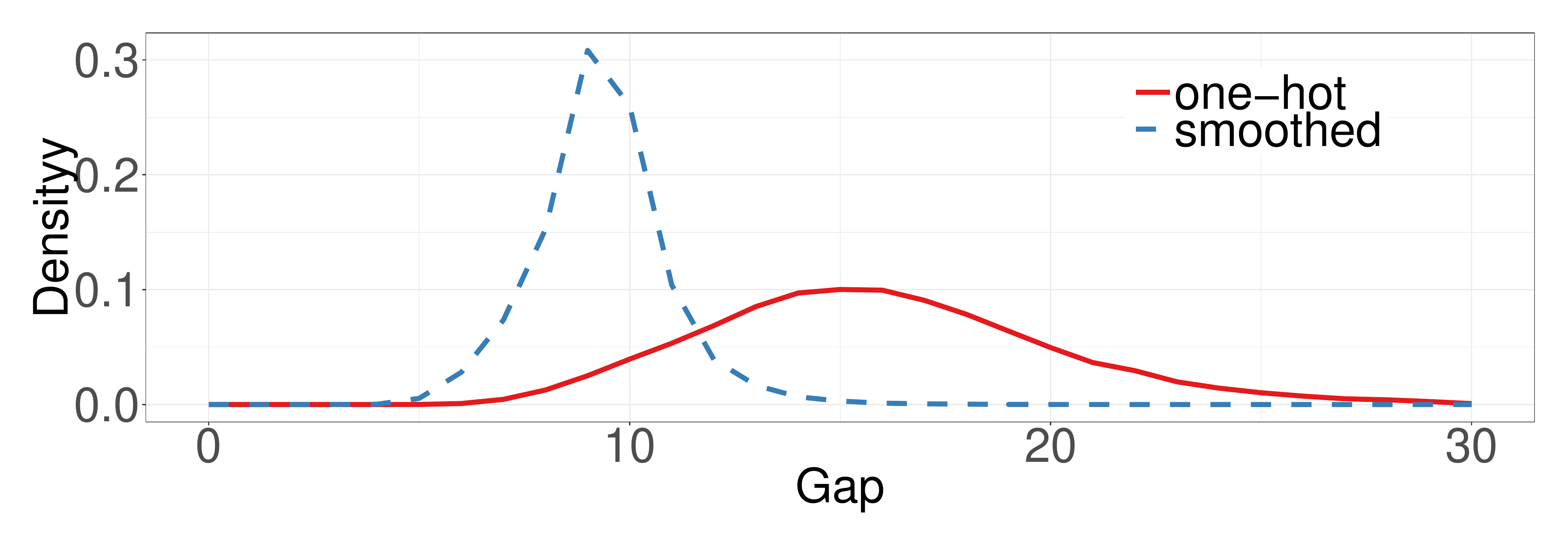}
    }%
  \caption{Visualization of the effectiveness of label smoothing on ImageNet. Top: theoretical gap between $z_p^*$ and others decreases when increasing
    $\varepsilon$. Bottom: The empirical distributions of the gap between the maximum prediction and the average of the rest.}
  \label{fig:learning-rate-curve}
\end{figure}

When $\varepsilon = 0$, the gap $\log( (K-1)(1-\varepsilon)/\varepsilon)$ will be $\infty$ and 
as $\varepsilon$ increases, the gap decreases. Specifically when $\varepsilon=(K-1)/K$, 
all optimal $z_i^*$ will be identical. Figure~\ref{fig:label_smoothing} shows how the gap
changes as we move $\varepsilon$, given $K=1000$ for ImageNet dataset.

We empirically compare the output value from two ResNet-50-D models that are trained with and without 
label smoothing respectively and calculate the gap between the maximum prediction value and 
the average of the rest. Under $\varepsilon=0.1$ and $K=1000$, the theoretical gap is 
around 9.1. Figure~\ref{fig:label_smoothing_density} demonstrate the gap distributions from the two models predicting
over the validation set of ImageNet. It is clear that with label smoothing the distribution centers
at the theoretical value and has fewer extreme values.

\subsection{Knowledge Distillation}

In knowledge distillation~\cite{hinton2015distilling}, we use a teacher model to
help train the current model, which is called the student model. The teacher
model is often a pre-trained model with higher accuracy, so by imitation, the
student model is able to improve its own accuracy while keeping the model
complexity the same. One example is using a ResNet-152 as the teacher model
to help training ResNet-50. 

During training, we add a distillation loss to penalize the difference between
the softmax outputs from the teacher model and the learner model. Given an input,
assume $p$ is the true probability distribution, and $z$ and $r$ are outputs of
the last fully-connected layer of the student model and the teacher model,
respectively. Remember previously we use a negative cross entropy loss $\ell(p, \textrm{softmax}(z))$
to measure the difference between $p$ and $z$, here we use the same loss again for the
distillation. Therefore, the loss is changed to

\begin{equation}
\ell(p, \textrm{softmax}(z)) + T^2 \ell(\textrm{softmax}(r/T), \textrm{softmax}(z/T)),
\end{equation}

where $T$ is the temperature hyper-parameter to make the softmax outputs
smoother thus distill the knowledge of label distribution from teacher's prediction.

\subsection{Mixup Training}

In Section~\ref{sec:basel-train-proc} we described how images are augmented
before training. Here we consider another augmentation method called
mixup~\cite{DBLP:journals/corr/abs-1710-09412}. In mixup, each time we randomly
sample two examples $(x_i, y_i)$ and $(x_j, y_j)$. Then we form a new example
by a weighted linear interpolation of these two examples:

\begin{eqnarray}
\hat x &=& \lambda x_i + (1-\lambda) x_j, \\
\hat y &=& \lambda y_i + (1-\lambda) y_j,
\end{eqnarray}

where $\lambda \in [0, 1]$ is a random number drawn from the
$\mathbf{Beta}(\alpha,\alpha)$ distribution. In
mixup training, we only use the new example $(\hat x, \hat y)$.

\subsection{Experiment Results}
\label{sec:adv}

\begin{table*}[t!]
  \centering
  \begin{tabular}{l|c|c|c|c|c|c}
    \hline
    \multirow{2}{*}{Refinements} &  \multicolumn{2}{c|}{ResNet-50-D} & \multicolumn{2}{c|}{Inception-V3} & \multicolumn{2}{c}{MobileNet} \\\cline{2-7}
    & Top-1 & Top-5 & Top-1 & Top-5 & Top-1 & Top-5 \\\specialrule{.1em}{.05em}{.05em}
    Efficient            & 77.16   & 93.52   &  77.50 & 93.60 & 71.90  & 90.53\\
    + cosine decay       & 77.91   & 93.81   &  78.19  & 94.06 &  72.83 & 91.00 \\
    + label smoothing    & 78.31   & 94.09   &   78.40 & 94.13   & 72.93  & 91.14 \\
    + distill w/o mixup  & 78.67   & 94.36   &   78.26 & 94.01   & 71.97  & 90.89 \\
    + mixup w/o distill  & 79.15   & 94.58   &  \textbf{78.77} &  \textbf{94.39} &  \textbf{73.28} & \textbf{91.30} \\
    + distill w/ mixup   & \textbf{79.29} & \textbf{94.63} & 78.34 & 94.16 & 72.51 &91.02 \\\hline
  \end{tabular}
  \caption{The validation accuracies on ImageNet for stacking training
    refinements one by one. The baseline models are obtained from Section~\ref{sec:efficient}.}
  \label{tab:train-refines}
\end{table*}

Now we evaluate the four training refinements. We set
$\varepsilon=0.1$ for label smoothing by
following Szegedy~\etal~\cite{DBLP:journals/corr/SzegedyVISW15}. For the model distillation we use $T=20$,
specifically a pretrained ResNet-152-D model with both cosine decay and label smoothing applied
is used as the teacher. In the mixup training, we choose
$\alpha=0.2$ in the Beta distribution and increase the number of epochs from 120
to 200 because the mixed examples ask for a longer training progress to converge better.
When combining the mixup training with distillation, we train the teacher model with mixup as well.

We demonstrate that the refinements are 
not only limited to ResNet architecture or the ImageNet dataset.
First, we train ResNet-50-D, Inception-V3 and MobileNet on ImageNet dataset with refinements.
The validation accuracies for applying these training refinements one-by-one are
shown in Table~\ref{tab:train-refines}. By stacking cosine decay, 
label smoothing and mixup, we have steadily improving ResNet, InceptionV3 and MobileNet models.
Distillation works well on ResNet,
however, it does not work well on Inception-V3 and MobileNet.
Our interpretation is that the teacher model is not from the same family of the student, therefore has
different distribution in the prediction, and brings negative impact to the model.

To support our tricks is transferable to other dataset, we train a ResNet-50-D model on MIT Places365 dataset with and without the refinements.
Results are reported in Table~\ref{tab:mitplaces}. We see the refinements improve
the top-5 accuracy consistently on both the validation and test set. 

\begin{table*}
\begin{center}
\begin{tabular}{l|c|c|c|c}
\hline
Model                   & Val Top-1 Acc    & Val Top-5 Acc  & Test Top-1 Acc  & Test Top-5 Acc\\ \specialrule{.1em}{.05em}{.05em}
ResNet-50-D Efficient   & 56.34          & 86.87        & 57.18         & 87.28 \\ \hline
ResNet-50-D Best        & \textbf{56.70}          & \textbf{87.33}        & \textbf{57.63}         & \textbf{87.82} \\ \hline
\end{tabular}
\end{center}
\caption{Results on both the validation set and the test set of MIT Places 365 dataset.
         Prediction are generated as stated in Section~\ref{sec:basel-train-proc}. ResNet-50-D Efficient refers to 
         ResNet-50-D trained with settings from Section~\ref{sec:efficient}, and ResNet-50-D Best further incorporate
         cosine scheduling, label smoothing and mixup. }
\label{tab:mitplaces}
\end{table*}

\section{Transfer Learning}
\label{sec:transfer-learning}

Transfer learning is one major down-streaming use case of trained image classification
models. In this section, we will investigate if these improvements discussed
so far can benefit transfer learning. In particular, we pick two important
computer vision tasks, object detection and semantic segmentation, and evaluate
their performance by varying base models.

\subsection{Object Detection}

\begin{table}
\begin{center}
\begin{tabular}{l|c|c}
\hline
Refinement   & Top-1 & mAP      \\ \specialrule{.1em}{.05em}{.05em}
B-standard & 76.14  & 77.54    \\
D-efficient  & 77.16    & 78.30     \\
+ cosine & 77.91  &  79.23\\
+ smooth       & 78.34     & 80.71   \\
+ distill w/o mixup  & 78.67   & 80.96   \\
+ mixup w/o distill  & 79.16 & 81.10   \\
+ distill w/ mixup &  79.29 & \textbf{81.33}  \\
\hline
\end{tabular}
\end{center}
\caption{Faster-RCNN performance with various pre-trained base networks evaluated on Pascal VOC.}
\label{tab:det}
\end{table}

The goal of object detection is to locate bounding boxes of objects in an
image. We evaluate performance using PASCAL VOC~\cite{pascal-voc-2007}. Similar to Ren~\etal~\cite{ren2015faster}, we use union
set of VOC 2007 \textit{trainval} and VOC 2012 \textit{trainval} for training, and VOC
2007 test for evaluation, respectively.  We train Faster-RCNN~\cite{ren2015faster} on
this dataset, with refinements from Detectron~\cite{Detectron2018} such as linear warmup and long training schedule.
The VGG-19 base model in Faster-RCNN is replaced with
various pretrained models in the previous discussion. We keep other settings the same so
the gain is solely from the base models.

Mean average precision (mAP) results are reported in Table~\ref{tab:det}. We
can observe that a base model with a higher validation accuracy leads to a higher
mAP for Faster-RNN in a consistent manner. In particular, the best base model with accuracy 79.29\% on ImageNet
leads to the best mAP at 81.33\% on VOC, which outperforms the standard model by 4\%.

\subsection{Semantic Segmentation}

\begin{table}
\begin{center}
\begin{tabular}{l|c|c|c}
\hline
Refinement   & Top-1      & PixAcc      & mIoU  \\ \specialrule{.1em}{.05em}{.05em}
B-standard  & 76.14    & 78.08     & 37.05 \\
D-efficient   & 77.16  & 78.88     & 38.88 \\
+ cosine    & 77.91    & \textbf{79.25}   & \textbf{39.33} \\
+ smooth   & 78.34     & 78.64     & 38.75 \\
+ distill w/o mixup & 78.67 & 78.97 & 38.90 \\
+ mixup w/o distill & 79.16 & 78.47 & 37.99 \\
+ mixup w/ distill & 79.29 & 78.72 & 38.40 \\
\hline
\end{tabular}
\end{center}
\caption{FCN performance with various base networks evaluated on ADE20K.}
\label{tab:seg}
\end{table}

Semantic segmentation predicts the category for every pixel from the input
images. We use Fully Convolutional Network (FCN)~\cite{long2015fully} for this task and train models on the
ADE20K~\cite{zhou2017scene} dataset.
Following PSPNet~\cite{zhao2017pyramid} and Zhang~\etal~\cite{Zhang_2018_CVPR}, we replace the base network
with various pre-trained models discussed in previous sections and apply dilation network
strategy~\cite{chen2018deeplab,yu2015multi} on stage-3 and stage-4. A fully
convolutional decoder is built on top of the base network to make the final
prediction.

Both pixel accuracy (pixAcc) and mean intersection over union (mIoU)
are reported in Table~\ref{tab:seg}.
In contradiction to our results on object detection, the cosine learning rate schedule
effectively improves the accuracy of the FCN performance, while other refinements
provide suboptimal results. A potential explanation to the phenomenon is that semantic segmentation predicts 
in the pixel level. While models trained with label smoothing, distillation and mixup
favor soften labels, blurred pixel-level information may be blurred and degrade overall pixel-level accuracy.

\section{Conclusion}

In this paper, we survey a dozen tricks to train deep convolutional neural
networks to improve model accuracy. These tricks introduce minor modifications to the model
architecture, data preprocessing, loss function, and
learning rate schedule. Our empirical results on ResNet-50, Inception-V3 and
MobileNet indicate that these tricks improve model accuracy consistently. More excitingly, stacking
all of them together leads to a significantly higher accuracy. In addition,
these improved pre-trained models show strong advantages in transfer learning, which improve both
object detection and semantic segmentation. We believe the benefits can extend to broader domains where classification base models are favored.

{\small
\bibliographystyle{ieee}
\bibliography{egbib}

\begin{thebibliography}{10}\itemsep=-1pt

\bibitem{DBLP:journals/corr/ChenPSA17}
L.~Chen, G.~Papandreou, F.~Schroff, and H.~Adam.
\newblock Rethinking atrous convolution for semantic image segmentation.
\newblock {\em CoRR}, abs/1706.05587, 2017.

\bibitem{chen2018deeplab}
L.-C. Chen, G.~Papandreou, I.~Kokkinos, K.~Murphy, and A.~L. Yuille.
\newblock Deeplab: Semantic image segmentation with deep convolutional nets,
  atrous convolution, and fully connected crfs.
\newblock {\em IEEE transactions on pattern analysis and machine intelligence},
  40(4):834--848, 2018.

\bibitem{pascal-voc-2007}
M.~Everingham, L.~Van~Gool, C.~K.~I. Williams, J.~Winn, and A.~Zisserman.
\newblock The {PASCAL} {V}isual {O}bject {C}lasses {C}hallenge 2007 {(VOC2007)}
  {R}esults.
\newblock
  http://www.pascal-network.org/challenges/VOC/voc2007/workshop/index.html.

\bibitem{ginsburg2018large}
B.~Ginsburg, I.~Gitman, and Y.~You.
\newblock Large batch training of convolutional networks with layer-wise
  adaptive rate scaling.
\newblock 2018.

\bibitem{Detectron2018}
R.~Girshick, I.~Radosavovic, G.~Gkioxari, P.~Doll\'{a}r, and K.~He.
\newblock Detectron.
\newblock \url{https://github.com/facebookresearch/detectron}, 2018.

\bibitem{glorot2010understanding}
X.~Glorot and Y.~Bengio.
\newblock Understanding the difficulty of training deep feedforward neural
  networks.
\newblock In {\em Proceedings of the thirteenth international conference on
  artificial intelligence and statistics}, pages 249--256, 2010.

\bibitem{DBLP:journals/corr/GoyalDGNWKTJH17}
P.~Goyal, P.~Doll{\'{a}}r, R.~B. Girshick, P.~Noordhuis, L.~Wesolowski,
  A.~Kyrola, A.~Tulloch, Y.~Jia, and K.~He.
\newblock Accurate, large minibatch {SGD:} training imagenet in 1 hour.
\newblock {\em CoRR}, abs/1706.02677, 2017.

\bibitem{resnet_torch}
S.~Gross and M.~Wilber.
\newblock Training and investigating residual nets.
\newblock http://torch.ch/blog/2016/02/04/resnets.html.

\bibitem{he2016deep}
K.~He, X.~Zhang, S.~Ren, and J.~Sun.
\newblock Deep residual learning for image recognition.
\newblock In {\em Proceedings of the IEEE conference on computer vision and
  pattern recognition}, pages 770--778, 2016.

\bibitem{hinton2015distilling}
G.~Hinton, O.~Vinyals, and J.~Dean.
\newblock Distilling the knowledge in a neural network.
\newblock {\em arXiv preprint arXiv:1503.02531}, 2015.

\bibitem{howard2017mobilenets}
A.~G. Howard, M.~Zhu, B.~Chen, D.~Kalenichenko, W.~Wang, T.~Weyand,
  M.~Andreetto, and H.~Adam.
\newblock Mobilenets: Efficient convolutional neural networks for mobile vision
  applications.
\newblock {\em arXiv preprint arXiv:1704.04861}, 2017.

\bibitem{hu2017squeeze}
J.~Hu, L.~Shen, and G.~Sun.
\newblock Squeeze-and-excitation networks.
\newblock {\em arXiv preprint arXiv:1709.01507}, 7, 2017.

\bibitem{huang2017densely}
G.~Huang, Z.~Liu, L.~van~der Maaten, and K.~Q. Weinberger.
\newblock Densely connected convolutional networks.
\newblock In {\em 2017 IEEE Conference on Computer Vision and Pattern
  Recognition (CVPR)}, pages 2261--2269. IEEE, 2017.

\bibitem{jia2018highly}
X.~Jia, S.~Song, W.~He, Y.~Wang, H.~Rong, F.~Zhou, L.~Xie, Z.~Guo, Y.~Yang,
  L.~Yu, et~al.
\newblock Highly scalable deep learning training system with mixed-precision:
  Training imagenet in four minutes.
\newblock {\em arXiv preprint arXiv:1807.11205}, 2018.

\bibitem{krizhevsky2012imagenet}
A.~Krizhevsky, I.~Sutskever, and G.~E. Hinton.
\newblock Imagenet classification with deep convolutional neural networks.
\newblock In {\em Advances in neural information processing systems}, pages
  1097--1105, 2012.

\bibitem{lin2013network}
M.~Lin, Q.~Chen, and S.~Yan.
\newblock Network in network.
\newblock {\em arXiv preprint arXiv:1312.4400}, 2013.

\bibitem{long2015fully}
J.~Long, E.~Shelhamer, and T.~Darrell.
\newblock Fully convolutional networks for semantic segmentation.
\newblock In {\em Proceedings of the IEEE conference on computer vision and
  pattern recognition}, pages 3431--3440, 2015.

\bibitem{DBLP:journals/corr/LoshchilovH16a}
I.~Loshchilov and F.~Hutter.
\newblock {SGDR:} stochastic gradient descent with restarts.
\newblock {\em CoRR}, abs/1608.03983, 2016.

\bibitem{micikevicius2017mixed}
P.~Micikevicius, S.~Narang, J.~Alben, G.~Diamos, E.~Elsen, D.~Garcia,
  B.~Ginsburg, M.~Houston, O.~Kuchaev, G.~Venkatesh, et~al.
\newblock Mixed precision training.
\newblock {\em arXiv preprint arXiv:1710.03740}, 2017.

\bibitem{nesterov1983method}
Y.~E. Nesterov.
\newblock A method for solving the convex programming problem with convergence
  rate o (1/k\^{} 2).
\newblock In {\em Dokl. Akad. Nauk SSSR}, volume 269, pages 543--547, 1983.

\bibitem{shufflenetv2}
H.-T.~Z. Ningning~Ma, Xiangyu~Zhang and J.~Sun.
\newblock Shufflenet v2: Practical guidelines for efficient cnn architecture
  design.
\newblock {\em arXiv preprint arXiv:1807.11164}, 2018.

\bibitem{ren2015faster}
S.~Ren, K.~He, R.~Girshick, and J.~Sun.
\newblock Faster r-cnn: Towards real-time object detection with region proposal
  networks.
\newblock In {\em Advances in neural information processing systems}, pages
  91--99, 2015.

\bibitem{russakovsky2015imagenet}
O.~Russakovsky, J.~Deng, H.~Su, J.~Krause, S.~Satheesh, S.~Ma, Z.~Huang,
  A.~Karpathy, A.~Khosla, M.~Bernstein, et~al.
\newblock Imagenet large scale visual recognition challenge.
\newblock {\em International Journal of Computer Vision}, 115(3):211--252,
  2015.

\bibitem{DBLP:journals/corr/SimonyanZ14a}
K.~Simonyan and A.~Zisserman.
\newblock Very deep convolutional networks for large-scale image recognition.
\newblock {\em CoRR}, abs/1409.1556, 2014.

\bibitem{smith2017don}
S.~L. Smith, P.-J. Kindermans, C.~Ying, and Q.~V. Le.
\newblock Don't decay the learning rate, increase the batch size.
\newblock {\em arXiv preprint arXiv:1711.00489}, 2017.

\bibitem{DBLP:journals/corr/SzegedyVISW15}
C.~Szegedy, V.~Vanhoucke, S.~Ioffe, J.~Shlens, and Z.~Wojna.
\newblock Rethinking the inception architecture for computer vision.
\newblock {\em CoRR}, abs/1512.00567, 2015.

\bibitem{xie2017aggregated}
S.~Xie, R.~Girshick, P.~Doll{\'a}r, Z.~Tu, and K.~He.
\newblock Aggregated residual transformations for deep neural networks.
\newblock In {\em Computer Vision and Pattern Recognition (CVPR), 2017 IEEE
  Conference on}, pages 5987--5995. IEEE, 2017.

\bibitem{yu2015multi}
F.~Yu and V.~Koltun.
\newblock Multi-scale context aggregation by dilated convolutions.
\newblock {\em arXiv preprint arXiv:1511.07122}, 2015.

\bibitem{DBLP:journals/corr/abs-1710-09412}
H.~Zhang, M.~Ciss{\'{e}}, Y.~N. Dauphin, and D.~Lopez{-}Paz.
\newblock mixup: Beyond empirical risk minimization.
\newblock {\em CoRR}, abs/1710.09412, 2017.

\bibitem{Zhang_2018_CVPR}
H.~Zhang, K.~Dana, J.~Shi, Z.~Zhang, X.~Wang, A.~Tyagi, and A.~Agrawal.
\newblock Context encoding for semantic segmentation.
\newblock In {\em The IEEE Conference on Computer Vision and Pattern
  Recognition (CVPR)}, June 2018.

\bibitem{zhao2017pyramid}
H.~Zhao, J.~Shi, X.~Qi, X.~Wang, and J.~Jia.
\newblock Pyramid scene parsing network.
\newblock In {\em Computer Vision and Pattern Recognition (CVPR), 2017 IEEE
  Conference on}, pages 6230--6239. IEEE, 2017.

\bibitem{zhou2017places}
B.~Zhou, A.~Lapedriza, A.~Khosla, A.~Oliva, and A.~Torralba.
\newblock Places: A 10 million image database for scene recognition.
\newblock {\em IEEE transactions on pattern analysis and machine intelligence},
  2017.

\bibitem{zhou2017scene}
B.~Zhou, H.~Zhao, X.~Puig, S.~Fidler, A.~Barriuso, and A.~Torralba.
\newblock Scene parsing through ade20k dataset.
\newblock In {\em Proceedings of the IEEE Conference on Computer Vision and
  Pattern Recognition}, 2017.

\bibitem{DBLP:journals/corr/ZophVSL17}
B.~Zoph, V.~Vasudevan, J.~Shlens, and Q.~V. Le.
\newblock Learning transferable architectures for scalable image recognition.
\newblock {\em CoRR}, abs/1707.07012, 2017.

\end{thebibliography}
}

\end{document}